\pdfoutput=1

\documentclass[11pt]{article}

\usepackage[final]{coling}

\usepackage{times,amsmath}
\usepackage{latexsym}
\usepackage{multirow}

\usepackage[T1]{fontenc}

\usepackage[utf8]{inputenc}

\usepackage{microtype}

\usepackage{inconsolata}

\usepackage{graphicx}

\title{Curriculum Learning for Cross-Lingual Data-to-Text Generation With Noisy Data}

\author{Kancharla Aditya Hari \\
  Sentisum \\
  \texttt{aditya@sentisum.com} \\\And
  Manish Gupta \\
  Microsoft,   Hyderabad, India \\
  \texttt{gmanish@microsoft.com}\\\And
  Vasudeva Varma \\
  IIIT Hyderabad \\
  \texttt{vv@iiit.ac.in} \\}

\begin{document}
\maketitle
\begin{abstract}
Curriculum learning has been used to improve the quality of text generation systems by ordering the training samples according to a particular schedule in various tasks. In the context of data-to-text generation (DTG), previous studies used various difficulty criteria to order the training samples for monolingual DTG. These criteria, however, do not generalize to the cross-lingual variant of the problem and do not account for noisy data. We explore multiple criteria that can be used for improving the performance of cross-lingual DTG systems with noisy data using two curriculum schedules. Using the alignment score criterion for ordering samples and an annealing schedule to train the model, we show increase in BLEU score by up to 4 points, and improvements in faithfulness and coverage of generations by 5-15\% on average across 11 Indian languages and English in 2 separate datasets. 
We make code and data publicly available\footnote{\url{https://tinyurl.com/yy5w8zu5}\label{footnote1}}.
\end{abstract}

\section{Introduction}
Data-to-text generation (DTG) is the task of transforming structured data, such as fact triples and tables into natural language \citep{Reiter_Dale_2000}. Cross-lingual DTG (XDTG) is a variant of the problem where the input data and the generated natural text are in different languages \citep{10.1145/3487553.3524265}. This is of particular relevance for low-resource languages 
as it allows input data from high resource languages such as English to be leveraged to generate text in low-resource languages \citep{cripwell-etal-2023-2023, singh2023xflt, sagare2023xf2t}. \citet{10.1145/3487553.3524265} created the \textsc{XAlign} dataset using automatic methods for XDTG from English to various Indian languages, the first such dataset for this task whereas the \textsc{WebNLG} 2023 Challenge \cite{cripwell-etal-2023-2023} focused on four under-resourced European languages. These datasets tend to be automatically curated, utilizing methods like machine translation.  

\begin{figure}
    \centering
    \includegraphics[width=0.5\textwidth]{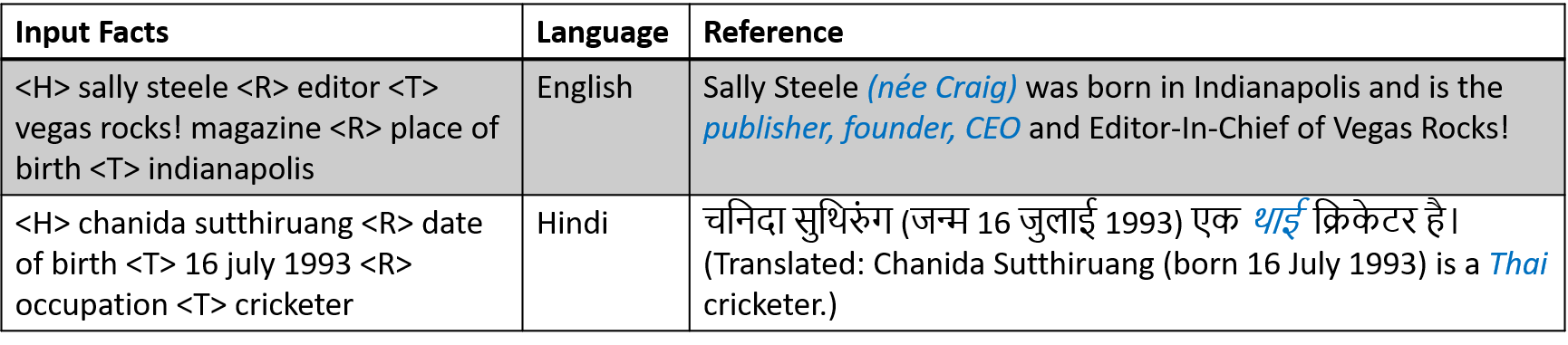}
    \caption{Examples of noisy data in the \textsc{XAlign} dataset~\cite{10.1145/3487553.3524265}. The input facts are represented as RDF triples with head, relation and tail tags (<H>, <R> and <T>). Highlighted text in the reference cannot be inferred from the input facts.}
    \label{fig:example}
\end{figure}

However, \citet{dhingra2019handling} observed that automatically curated datasets are prone to being noisy. A particular problem observed was that of divergent references where the reference texts deviate from the input data. This can manifest as either including information that cannot be inferred from the input data or omitting information that is present in the input data. Two examples of such noisy samples are shown in Fig.~\ref{fig:example}.

In this work, we use curriculum learning, where training samples are presented to the model in a specific simple-to-difficult order based on certain criteria, to improve performance of neural methods for XDTG. The approach has been shown to improve performance of monolingual DTG systems \cite{chang-etal-2021-order} using various difficulty criteria. The applicability of these approaches to XDTG however has not been studied as the cross-lingual setting presents unique challenges. Previous work has shown that criteria that jointly model both the input and the target perform better than criteria based on only one. However, such criteria, defined for monolingual DTG, cannot be easily adapted to the cross-lingual setting. Moreover, existing works only study schedules based on the notion of increasing difficulty which does not account for potential noise in the data. Curriculum learning with ``annealing'' data, that progressively removes examples of lower quality, has previously been used for learning from noisy data in other tasks \citep{wang-etal-2018-denoising, Hirsch_2024_WACV}, and its utility for the XDTG task bears further investigation, and is therefore the focus of our work. To demonstrate the potential of our method, we use the existing \textsc{XAlign} dataset, and also introduce a new cross-lingual noisy dataset based on the ToTTo dataset \cite{parikh-etal-2020-totto} called \textsc{xToTTo}. 

Overall, we make the following contributions in this paper. 
    (1) We propose the usage of curriculum learning for the XDTG problem with noisy input data.
    (2) We empirically study the behaviour of two curriculum learning schedules (expanding and annealing) with various ordering criteria on two XDTG datasets.  
    (3) We propose a new quality based cross-lingual criterion (alignment score) and show that with an annealing approach, it results in the best performance, evaluating performance with a combination of automatic metrics including LLM-based evaluation and human evaluation.
    (4) We make code and data publicly available\footref{footnote1}.

\section{Related Work}
Various neural approaches have been investigated for XDTG. \citet{10.1145/3487553.3524265} put forward the \textsc{XAlign} dataset and established baselines using seq2seq models. \citet{sagare2023xf2t} investigated multilingual pretraining and fact-aware embeddings, while \citet{singh2023xflt} focus on DTG for long text. \citet{10.1007/978-3-030-62419-4_24} used a graph attention network based encoder and a transformer decoder to verbalize RDF triples in English, German and Russian using the enriched version of the \textsc{WebNLG} dataset \citep{castro-ferreira-etal-2018-enriching}. 

\citet{10.1145/1553374.1553380} showed empirically that curriculum learning has an effect on both the convergence speed and, in some cases, the quality of local minima obtained. For text based tasks, n-gram frequency, token rarity and sentence length are some criteria used which are based only on the input or output text \citep{kocmi-bojar-2017-curriculum, platanios-etal-2019-competence, liu-etal-2020-norm, chang-etal-2021-order}. \citet{kocmi-bojar-2017-curriculum} also use linguistic features such as number of coordinating conjunctions. Criteria such as data uncertainty \citep{zhou-etal-2020-uncertainty}, 
and edit distances have been used to jointly consider both the input and output \citep{chang-etal-2021-order}. 
Some methods have also been designed for generating faithful text from noisy data, such as loss truncation \cite{kang-hashimoto-2020-improved} and controlled hallucinations \cite{filippova-2020-controlled}.

\section{Curriculum Learning Strategy}

\begin{figure}
    \centering
    \includegraphics[width=0.4\textwidth]{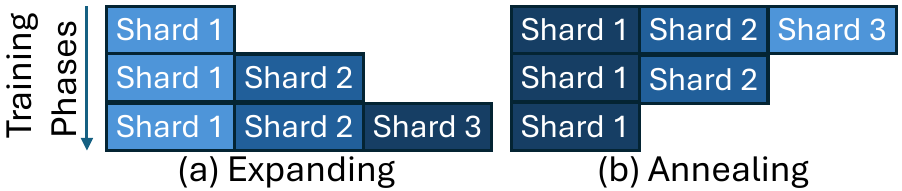}
    \caption{Two curriculum schedules: (a) Expanding schedule introduces new shards, (b) annealing schedule removes shards as the training progresses. Each row represents a training phase. 
    }
    \label{fig:curriculums}
\end{figure}

\noindent\textbf{Annealing vs Expanding Approach.} 
We use a probabilistic curriculum learning strategy similar to the one used by \citet{zhang2018empirical} for machine translation. The training samples are first distributed into distinct \textit{shards} based on the value of the chosen criterion. The training process is segmented into different \textit{phases}, with samples selected from only a subset of shards in a phase. We experiment with two approaches for selecting the shards, as shown in Fig.~\ref{fig:curriculums}. The first is to begin the training with the shard with the lowest scores, with shards added in the subsequent phases in ascending order. We term this the \textit{expanding} approach. The \textit{annealing} approach is based on studies related to learning from noisy data for other NLG tasks. Here, the training begins with every shard available in the first phase and shards with the lowest scores are removed in subsequent phases. The shards are shuffled within a specific phase. 
We choose this strategy over other curriculum learning strategies due to its flexible nature, requiring only a modification to the sampling strategy.

\noindent\textbf{Curriculum Criteria.} We consider two criteria used by \citet{chang-etal-2021-order} for monolingual DTG: sequence length and word rarity. Their soft edit distance criterion cannot be used for cross-lingual learning as it relies on exact sequence matching between the input and reference text.  Hence, we introduce a new cross-lingual criterion to jointly consider the input and target text: alignment. 

\noindent\textbf{(1) Length}: Generating longer sentences is more challenging, as errors made early in the decoding process propagate further. For a sequence $s = \{w_1, w_2 \ldots w_N\}$, 
$d_{\text{length}}(s)=N$.

\begin{table*}[!t]
\scriptsize
\centering
\begin{tabular}{|l|l|r|r|r|r|r|r|r|r||r|r|r|r|r|r|r|r|r|r|}
\hline
&\multirow{3}{*}{Lang}&\multicolumn{8}{c||}{BLEU}&\multicolumn{8}{c|}{chrf++}\\
\cline{3-18}
&&\multicolumn{1}{c|}{\multirow{2}{*}{Base}}&\multicolumn{1}{c|}{\multirow{2}{*}{LT}}&\multicolumn{2}{c|}{Length}&\multicolumn{2}{c|}{Rarity}&\multicolumn{2}{c||}{Alignment}&\multicolumn{1}{c|}{\multirow{2}{*}{Base}}&\multicolumn{1}{c|}{\multirow{2}{*}{LT}}&\multicolumn{2}{c|}{Length}&\multicolumn{2}{c|}{Rarity}&\multicolumn{2}{c|}{Alignment}\\
\cline{5-10}
\cline{13-18}
&&&&\multicolumn{1}{c|}{E}&\multicolumn{1}{c|}{A}&\multicolumn{1}{c|}{E}&\multicolumn{1}{c|}{A}&\multicolumn{1}{c|}{E}&\multicolumn{1}{c||}{A}&&&\multicolumn{1}{c|}{E}&\multicolumn{1}{c|}{A}&\multicolumn{1}{c|}{E}&\multicolumn{1}{c|}{A}&\multicolumn{1}{c|}{E}&\multicolumn{1}{c|}{A}\\
\hline
\multirow{13}{*}{\rotatebox{90}{\textsc{XAlign}}}&as&\textbf{12.60}&8.14&10.26&8.22&8.19&7.20&7.59&11.53&\textbf{34.81}&26.54&31.99&28.33&29.04&29.50&29.17&33.43\\
&bn&48.03&44.33&43.13&\textbf{68.05}&38.65&67.76&42.78&64.78&74.30&67.99&68.40&42.87&82.10&81.89&67.47&\textbf{83.86}\\
&en&46.39&41.33&45.42&42.20&46.06&43.45&39.59&\textbf{48.63}&64.37&59.29&59.54&55.84&60.80&56.93&55.91&\textbf{64.41}\\
&gu&21.70&10.45&18.11&21.43&19.11&22.23&19.13&\textbf{22.83}&49.24&37.29&45.46&45.99&45.57&47.23&43.56&\textbf{51.17}\\
&hi&41.75&36.29&41.82&42.80&38.88&\textbf{47.43}&39.90&46.41&66.46&60.70&64.76&63.66&62.96&65.96&63.68&\textbf{67.17}\\
&kn&9.14&3.95&7.37&10.66&7.69&12.14&7.82&\textbf{12.70}&45.09&32.24&39.32&40.28&39.10&41.54&38.51&\textbf{46.58}\\
&ml&25.71&22.14&25.38&24.91&25.08&26.21&24.57&\textbf{28.50}&56.26&49.24&53.97&52.20&52.09&52.75&50.32&\textbf{56.94}\\
&mr&24.81&15.09&25.31&27.23&23.62&27.98&25.42&\textbf{28.90}&56.66&46.34&52.06&50.89&50.56&52.11&49.88&\textbf{58.35}\\
&or&\textbf{44.09}&31.68&32.15&26.90&36.23&32.00&35.30&42.93&\textbf{68.19}&53.11&57.80&49.28&58.71&53.70&58.61&64.61\\
&pa&25.71&13.92&19.29&27.70&21.46&27.44&23.65&\textbf{29.25}&50.61&38.51&44.38&46.06&44.68&46.57&44.55&\textbf{52.16}\\
&ta&18.92&8.97&16.83&19.66&13.02&18.93&21.25&\textbf{22.22}&54.12&43.46&52.16&51.81&46.42&54.22&53.36&\textbf{59.12}\\
&te &13.34&6.64&10.70&13.27&11.21&14.37&10.45&\textbf{15.56}&50.03&38.90&44.90&44.64&43.69&46.33&42.95&\textbf{50.99}\\
\cline{2-18}
&Avg&27.06&20.05&24.61&28.60&23.56&29.65&24.59&\textbf{31.18}&55.77&46.45&51.52&51.95&50.19&53.29&50.05&\textbf{57.77}\\
\hline
\hline
\multirow{13}{*}{\rotatebox{90}{\textsc{xToTTo}}}&as&19.11&10.26&16.77&20.03&17.38&19.76&15.68&\textbf{20.24}&46.11&32.36&44.85&41.44&44.75&43.55&43.16&\textbf{47.82}\\
&bn&19.03&11.60&17.69&19.65&18.16&19.89&16.70&\textbf{20.70}&49.10&35.80&47.80&44.38&47.86&46.29&46.74&\textbf{50.16}\\
&en&27.33&16.61&27.46&29.43&28.86&29.51&26.28&\textbf{32.43}&50.01&30.35&49.58&46.75&51.14&48.61&47.85&\textbf{51.94}\\
&gu&21.86&13.44&21.56&24.05&22.00&24.14&18.77&\textbf{25.14}&47.78&32.85&46.29&43.15&46.54&44.63&44.21&\textbf{48.13}\\
&hi&23.32&13.08&21.70&24.35&23.14&25.35&19.88&\textbf{25.87}&47.66&32.61&46.59&43.42&47.15&45.24&44.63&\textbf{48.62}\\
&kn&20.05&9.17&18.45&19.37&19.73&20.71&16.84&\textbf{20.97}&50.24&34.84&48.50&45.49&48.88&47.48&47.11&\textbf{51.36}\\
&ml&18.00&7.53&15.84&18.31&16.73&19.19&14.78&\textbf{19.82}&45.98&28.36&43.40&41.38&44.20&42.87&42.13&\textbf{46.57}\\
&mr&18.65&9.75&18.02&20.32&17.96&\textbf{21.44}&16.69&21.23&45.84&30.58&44.01&40.94&44.35&42.80&43.43&\textbf{46.31}\\
&or&15.99&9.59&15.59&17.54&14.35&\textbf{17.62}&13.78&17.23&44.88&31.94&43.22&41.22&42.88&41.94&42.15&\textbf{45.60}\\
&ta&21.04&10.82&18.56&22.53&19.99&21.72&16.58&\textbf{22.90}&49.10&34.40&47.01&45.59&47.86&46.60&45.48&\textbf{50.31}\\
&te &18.90&8.61&17.09&19.45&17.43&19.75&15.13&\textbf{20.75}&49.30&32.29&46.93&44.67&47.08&46.19&45.16&\textbf{49.88}\\
\cline{2-18}
&Avg&20.30&10.95&18.98&21.37&19.61&21.73&17.37&\textbf{22.48}&47.82&32.40&46.20&43.49&46.61&45.11&44.73&\textbf{48.79}\\
\hline
\end{tabular}
\caption{BLEU and chrf++ scores for \textsc{XAlign} and \textsc{xToTTo} datasets using models trained without curriculum learning - baseline method (base) and loss truncation (LT), and trained using curriculum learning with sequence length, word rarity and alignment criteria with expanding (E) and annealing (A) schedules.}
\label{tab:XAlign-xToTTo-chrf-bleu}
\end{table*}

\noindent\textbf{(2) Rarity}: This is the sequence probability using a unigram model.  Based on the intuition that rarer words are harder to generate, rarity implicitly encodes information about the frequency of the words in the sequence, as well as the sequence length. $d_{\text{rarity}}(s)=-\sum_{k=1}^{N}\log{p(w_k)}$ where the unigram probability of $w_i$ is given as $p(w_i)$.

\noindent\textbf{(3) Alignment}: \citet{singh2023xflt} and \citet{filippova-2020-controlled} showed that quantifying the alignment between input facts and reference texts can be used to improve the quality of generation in XDTG when the reference text is partially aligned. They train a binary classifier to classify fact-text pairs as having complete or partial alignment using a small, manually annotated dataset. Labels are then assigned to each example based on confidence scores of this classifier.. 
We propose using this confidence score as the criterion for ordering the samples. We call this criterion the alignment score. For \textsc{XAlign}, we train a MURIL model~\citep{khanuja2021muril} using data and code from \citet{singh2023xflt}\footnote{\url{https://github.com/bhavyajeet/XFLT}}. 
For \textsc{xToTTo}, we use GPT-4~\citep{openai2023gpt} to annotate samples for partial and complete alignment.

\section{Experimental Setup}
We perform experiments on two datasets: \textsc{XAlign} and \textsc{xToTTo}. \textsc{XAlign} has 0.45M cross-lingual fact-text pairs in 13 languages: English (en), and 12 Indian languages: Assamese   (as), Bangla   (bn), Gujarati  (gu), Hindi     (hi), Kannada     (kn), Malayalam     (ml), Marathi   (mr), Odia   (or), Punjabi   (pa), Tamil     (ta), Telugu  (te). \textsc{XAlign} was automatically generated by aligning facts represented as RDF triples from Wikidata to sentences from Wikipedia using transfer learning. This results in partially aligned data containing noisy samples. Test set has 5042 manually annotated samples.

We also construct a new XDTG dataset based on the ToTTo dataset called \textsc{xToTTo}. We use the round-trip translation strategy used by \citet{zhu-etal-2019-ncls}. Given a data-text pair $(D, T)$, we first translate $T$ into the target language $L$, and then back to English to obtain $\hat{T}$. Then, cross-lingual pairs with    ROUGE1$(T, \hat{T})>R_1$ and ROUGE2$(T, \hat{T})>R_2$ are selected. We use the 3.3B NLLB model \cite{nllb2022} to translate the texts in the \textsc{XAlign} languages except Punjabi, which NLLB does not support.   
Since \textsc{xToTTo} is based on the noisy annotations in the ToTTo dataset, \textsc{xToTTo} also inherits the noise. We report scores on the validation set as the test set is hidden. \textsc{xToTTo} has $\sim$817K train and $\sim$90K validation samples.

\begin{table*}[!t]
\scriptsize
\centering
\begin{tabular}{|l|lll|lll|lll||lll|lll|lll|}
\hline
 &\multicolumn{9}{c||}{\textsc{XAlign}}&\multicolumn{9}{c|}{\textsc{xToTTo}}\\
 \hline
&\multicolumn{3}{c|}{Loss Truncation}&\multicolumn{3}{c|}{Baseline}&\multicolumn{3}{c||}{Alignment (\textsc{Alg-Ann})}&\multicolumn{3}{c|}{Loss Truncation}&\multicolumn{3}{c|}{Baseline}&\multicolumn{3}{c|}{Alignment (\textsc{Alg-Ann})}\\
\hline
&Fl.&Fa.&Cov.&Fl.&Fa.&Cov.&Fl.&Fa.&Cov.&Fl.&Fa.&Cov.&Fl.&Fa.&Cov.&Fl.&Fa.&Cov.\\
\hline
as&0.60&0.25&1.44&\textbf{0.86}&0.28&1.64&0.83&\textbf{0.38}&\textbf{1.84}       &0.70&0.16&2.34&0.93&0.39&\textbf{2.56}&\textbf{0.95}&\textbf{0.40}&2.52\\
bn&0.86&0.41&1.46&\textbf{0.95}&0.51&1.43&0.90&\textbf{0.67}&\textbf{1.60}       &0.63&0.16&2.46&0.92&0.35&\textbf{2.60}&\textbf{0.94}&\textbf{0.38}&2.58\\
en&0.76&0.23&2.41&\textbf{0.89}&0.45&2.50&0.87&\textbf{0.55}&\textbf{2.72}       &0.54&0.24&2.33&\textbf{0.94}&0.38&\textbf{2.53}&0.93&\textbf{0.43}&\textbf{2.53}\\
gu&0.81&0.37&1.83&0.81&0.45&1.95&\textbf{0.87}&\textbf{0.66}&\textbf{2.23}       &0.78&0.22&2.48&0.94&0.34&2.58&\textbf{0.96}&\textbf{0.38}&\textbf{2.59}\\
hi&0.88&0.53&1.96&\textbf{0.95}&0.52&2.01&0.94&\textbf{0.63}&\textbf{2.03}       &0.68&0.17&2.19&0.93&0.31&\textbf{2.56}&\textbf{0.95}&\textbf{0.34}&2.55\\
kn&0.72&0.33&1.76&\textbf{0.78}&0.39&1.82&0.77&\textbf{0.54}&\textbf{2.30}       &0.68&0.13&2.39&0.92&0.36&2.57&\textbf{0.96}&\textbf{0.38}&\textbf{2.59}\\
ml&0.82&0.52&1.58&0.94&0.47&1.72&\textbf{0.98}&\textbf{0.70}&\textbf{1.87}       &0.58&0.14&2.25&0.89&0.32&2.53&\textbf{0.92}&\textbf{0.34}&\textbf{2.55}\\
mr&0.70&0.38&1.85&0.88&0.41&1.98&\textbf{0.89}&\textbf{0.55}&\textbf{2.13}       &0.69&0.20&2.34&0.93&0.33&\textbf{2.56}&\textbf{0.95}&\textbf{0.37}&2.55\\
or&0.70&0.38&2.19&0.94&0.30&2.26&\textbf{0.97}&\textbf{0.52}&\textbf{2.53}       &0.67&0.21&2.48&0.90&0.43&2.54&\textbf{0.93}&\textbf{0.47}&\textbf{2.56}\\
pa&0.75&0.34&1.96&\textbf{0.77}&0.40&2.07&0.74&\textbf{0.55}&\textbf{2.40}       &\multicolumn{9}{c|}{----- ----- \textsc{xToTTo} does not have Punjabi (pa) samples.----- ----- }\\
ta&0.78&0.40&1.65&\textbf{0.87}&0.49&1.75&0.85&\textbf{0.66}&\textbf{1.80}       &0.52&0.12&2.32&0.90&0.28&\textbf{2.52}&\textbf{0.92}&\textbf{0.30}&2.50\\
te&0.74&0.36&1.83&0.83&0.36&1.88&\textbf{0.87}&\textbf{0.54}&\textbf{2.17}       &0.63&0.18&2.19&0.90&0.32&\textbf{2.55}&\textbf{0.93}&\textbf{0.35}&\textbf{2.55}\\
\hline
Avg&0.76&0.38&1.79&\textbf{0.87}&0.43&1.88&\textbf{0.87}&\textbf{0.58}&\textbf{2.08}&0.65&0.18&2.34&0.92&0.34&\textbf{2.55}&\textbf{0.94}&\textbf{0.38}&\textbf{2.55}\\
\hline
\end{tabular}
\caption{Fluency, faithfulness and coverage 
using GPT4-based evaluation. See Appendix~\ref{app:examples} for example outputs.}
\label{tab:XAlign-xtotto-llm-eval}
\end{table*}

For every curriculum criterion, we performed experiments with both expanding as well as annealing schedule. We train baseline models without a curriculum learning strategy, and also compare the performance of our proposed approach with loss truncation \cite{kang-hashimoto-2020-improved}. This method involves dropping examples that have losses above a certain quantile estimate during gradient descent.

We train mT5~\citep{xue-etal-2021-mt5}-small models with 300M parameters. Appendix~\ref{app:hyperparams} provides detailed hyper-parameter settings. 
\section{Results and Analysis}

\noindent\textbf{Standard Generation Metrics.} 
Table \ref{tab:XAlign-xToTTo-chrf-bleu} shows that for both datasets, an annealing schedule with alignment criterion (\textsc{Alg-Ann}) results in the model with the highest BLEU and chrf++ scores. However, using length and rarity for ordering samples results in worse performance than the baseline method without curriculum learning. Perhaps, this is due to the inability of these criteria to deal with noisy data. Loss truncation leads to lowest scores, indicating its inefficacy in dealing with multilingual data. 
For both datasets and all ordering criteria, annealing schedule results in better scores than expanding schedule. While curriculum learning typically relies on the assumption that the performance of the model increases if ``difficult'' data is slowly added during training, the trend suggests that with noisy data it is important to refine training with highest quality data as the training progresses.  

\noindent\textbf{LLM Evaluation.}
Traditional automatic metrics are inadequate to measure nuanced aspects which are important for XDTG evaluation. Recent works have shown that LLM-based evaluation is consistent with expert human evaluation \cite{chiang-lee-2023-large}. Hence, we use GPT-4 to evaluate our proposed method across three parameters: Fluency (Fl.), Faithfulness (Fa.) and Coverage (Cov.). Fl. and Fa. are on a scale of 0-1. Cov measures number of covered facts. Detailed prompts are in Appendix~\ref{app:prompts}.
Table~\ref{tab:XAlign-xtotto-llm-eval} compares best curriculum learning model (\textsc{Alg-Ann}) with baseline and loss truncation methods. Both  baseline and \textsc{Alg-Ann} result in fluent text. 
However, \textsc{Alg-Ann} results in higher faithfulness and coverage. In \textsc{XAlign}, it results in $15\%$ absolute increase in Fa. and $0.4$ higher Cov. The difference in less pronounced in \textsc{xToTTo}, with $4\%$ absolute increase in Fa. but similar Cov. While \textsc{XAlign} was annotated by humans for alignment, GPT-4 was used to annotate \textsc{xToTTo} which could be responsible for the narrower improvement. Further, \textsc{xToTTo} requires greater reasoning over cells, making the task more challenging than \textsc{XAlign}.

\begin{table}[!t]
\scriptsize
\centering
\begin{tabular}{|l|l|l|l|l||l|l|l|}
\hline
       && \multicolumn{3}{c||}{Baseline}                                                 & \multicolumn{3}{c|}{\textsc{Alg-Ann}}                                              \\ \hline
       && \multicolumn{1}{l|}{Fl.} & \multicolumn{1}{l|}{Fa.} & Cov. & \multicolumn{1}{l|}{Fl.} & \multicolumn{1}{l|}{Fa.} & Cov. \\ \hline
\multirow{3}{*}{\textsc{XAlign}} 
&en&0.84&0.51&2.55&\textbf{0.86}&\textbf{0.64}&\textbf{2.61} \\ 
&hi&\textbf{0.82}&0.55&1.92&0.81&\textbf{0.70}&\textbf{1.93} \\ 
&te&0.78&0.53&1.42&\textbf{0.80}&\textbf{0.68}&\textbf{1.57} \\ \hline
\multirow{3}{*}{\textsc{xToTTo}} 
&en&0.87&0.53&2.50&\textbf{0.88}&\textbf{0.60}&\textbf{2.52} \\ 
&hi&0.76&0.42&2.21&\textbf{0.78}&\textbf{0.47}&\textbf{2.24} \\ 
&te&0.72&0.36&2.13&\textbf{0.77}&\textbf{0.42}&\textbf{2.33} \\ \hline
\end{tabular}
\caption{Human evaluation results.}
\label{tab:human-eval}
\end{table}

\noindent\textbf{Human Evaluation.} 100 sample generations of our \textsc{Alg-Ann} method and the baseline are annotated by 3 annotators for en, hi and te due to availability constraints for expert annotators for other langs. Table~\ref{tab:human-eval} shows that \textsc{Alg-Ann} outperforms baseline in all three metrics. In \textsc{XAlign}, the average faithfulness of generated texts increases by $14\%$, while in \textsc{xToTTo} it increases by $6\%$. Human evaluation results also reveal that LLM-based evaluation overestimates the coverage of the generated texts, while underestimating their faithfulness. The higher coverage can be explained by the inability of GPT-4 to accurately extract the facts from the text, especially for multilingual data, while also often counting incorrectly verbalized data. Further, GPT-4 is stricter than human evaluators at marking examples as containing unsupported information. See Appendix~\ref{app:examples} for example outputs. 

\section{Conclusion}  
We show that using curriculum learning with noisy data in a cross-lingual setting results in promising improvements in quality of XDTG. We show that standard difficulty based criteria are not suited for this task. Instead, our novel quality based criterion (alignment) when combined with an annealing schedule where as the training progresses the model is exposed to only the highest quality data results in the best performing model. An increase in automatic metrics (BLEU and chrf++) is observed, along with improvement in quality attributes as measured by both LLMs and human evaluation. The generated text is more fluent, is more faithful, and covers a greater amount of the input facts. 

\section{Limitations} 
The proposed alignment criterion requires annotating data-text pairs for alignment. While we demonstrate its potential when obtained using an LLM, the improvement in performance is smaller than that obtained using human annotations.

Further, while we investigate the problem of data-to-text generation, several other tasks such as headline generation, abstractive summarization etc. require text-grounded generation. These could also benefit from the method, and would establish its generalizability to other tasks. 

\section{Ethical Considerations}
While our method considers the noisy nature of the data, the risk of generating hallucinatory text exists as with any other neural NLG system. Applying ideas from works focusing on reducing hallucinations could benefit the system. 

The \textsc{XAlign} V2 dataset is released under the MIT license on GitHub\footnote{\url{https://github.com/tushar117/XAlign}}. The dataset contains 12 languages, which are shown in Appendix Table \ref{tab:langcodes}. The \textsc{ToTTo} dataset is released under the Creative Commons Share-Alike 3.0 license on Github\footnote{\url{https://github.com/google-research-datasets/ToTTo}}.

mT5-small model\footnote{\url{https://huggingface.co/google/mt5-small}} is also publicly released under Apache License 2.0. 

\bibliography{main}

\newpage

\appendix

\section{Language Codes}
\label{lang_codes}
The language codes are shown in Table \ref{tab:langcodes}.

\begin{table}[h!]
\centering
\scriptsize
\begin{tabular}{|l|l|}
\hline
Language Code           & Language \\ \hline
as & Assamese  \\
bn & Bangla  \\
en & English  \\
gu & Gujarati \\
hi & Hindi    \\
kn & Kannada    \\
ml & Malayalam    \\
mr & Marathi  \\
or & Odia  \\
pa & Punjabi  \\
ta & Tamil    \\
te & Telugu   \\ \hline
\end{tabular}
\caption{Language codes of the languages included in the \textsc{XAlign} and \textsc{xToTTo} datasets}
\label{tab:langcodes}
\end{table}

\section{Hyper-parameters for Reproducibility}
\label{app:hyperparams}
We train mT5~\citep{xue-etal-2021-mt5}-small models with 300M parameters. 
We use Adafactor optimizer with an initial learning rate of 0.001, with a linear decaying schedule. For curriculum learning, the data was divided into 8 shards for all languages. For all experiments, the model with the lowest validation loss was picked. Also, we used ROUGE thresholds $R_1=0.70$ and $R_2=0.35$. 

\section{LLM Prompts}
\label{app:prompts}
\subsection{Alignment annotation prompt}
*** OBJECTIVE *** \\ 
You are provided with data and the corresponding verbalization of the data. The data is from \textsc{Name} dataset which is a data-to-text dataset.
The texts are in \#lang\# and the data in English. 
Your task is determine if the alignment between the data and the text is PARTIAL or COMPLETE. 
PARTIAL alignment means that the text is not faithful to the input text and it contains extra information that is not present in the data. 
COMPLETE alignment means that the table is sufficient to infer the text, and the text contains no extra information. Focus on the meaning of the text rather than exact phrasing or verbiage; the task may require reasoning over the data, so not everything needs to be explicitly verbalized. 

\subsection{LLM-evaluation prompt}
*** OBJECTIVE *** \\ 
You are an expert for evaluating the quality of outputs generated by an NLG system for data-to-text generation. In this task, you will be given a pair of input data and the corresponding output for this. The data is from the ToTTo dataset, and in the form of a subtable, with the contents of the cell and its column header provided as well the title of the page and the section it belongs to on Wikipedia. You are required to evaluate the response on the following three parameters -  

1. FLUENCY 
The fluency is a the measure of how natural and grammatically correct the text is. Choose one of three options - "fluent", "mostly fluent", "not fluent"  

2. FAITHFULNESS 
This is a measure of how the level of hallucination in the text with respect to the table. The text should express only information supported by the table or by non-expert background knowledge. 
- "faithful" - if the text has NO hallucinations or unsupported information and requires no expert background knowledge, 
- "mostly faithful" - if the text contains NO hallucinations or unsupported information but requires some background knowledge 
- "not faithful" - if the text contains ANY hallucinations or unsupported information 
Further, produce the list of phrases that are hallucinations/unsupported.  

3. COVERAGE 
This is a measure of how much of the provided data is expressed in the input text. For this, count the cells that the generated text covers. Note that the text does not need to explicitly verbalize the information, and may be in the form of reasoning over the cells. 

\section{Examples}
\label{app:examples}
\begin{figure*}
    \centering
    \includegraphics[width=\textwidth]{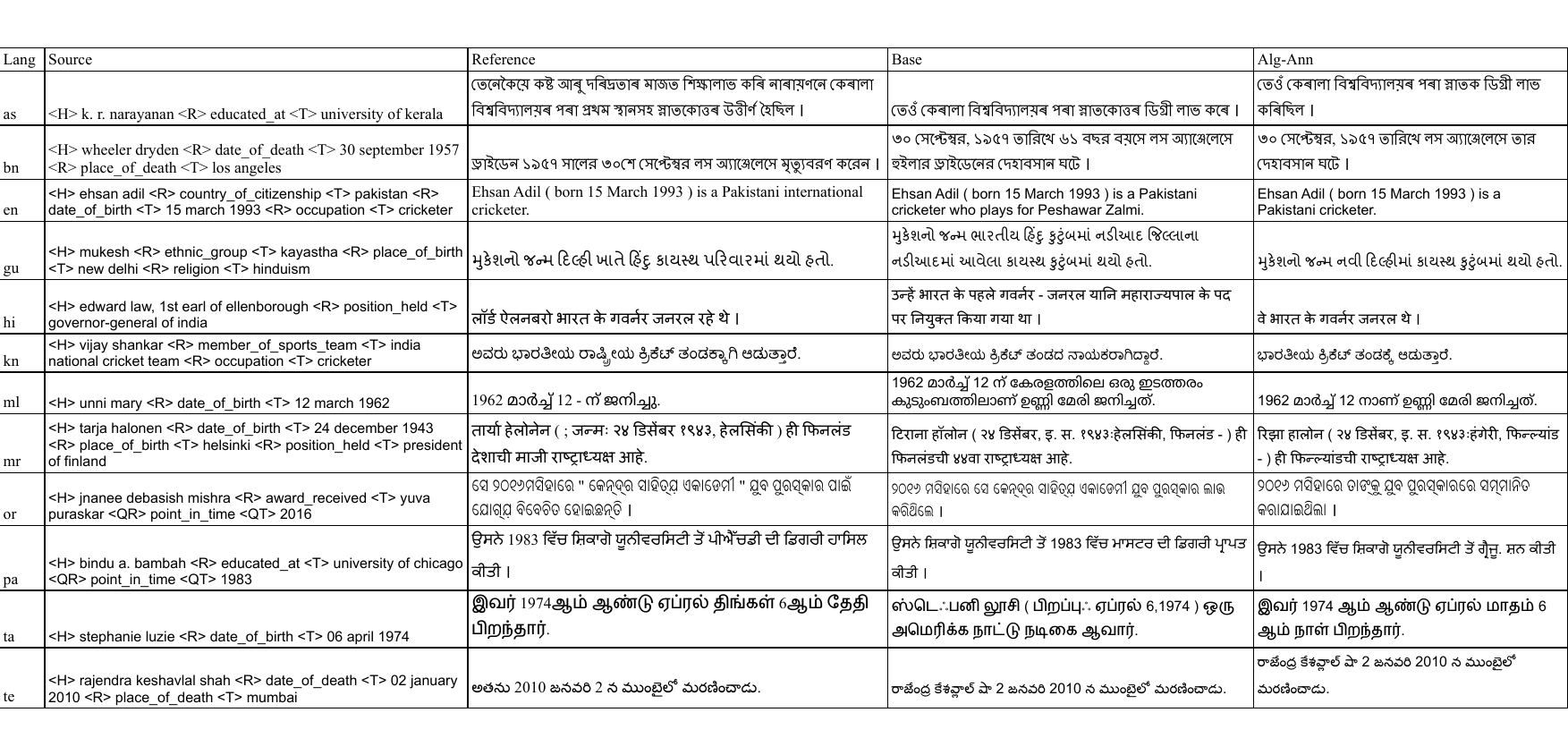}
    \captionof{table}{Some examples of generation from \textsc{XAlign
    } using the best performing model compared to baseline model}
    \label{fig:xalign_examples}
\end{figure*}

\begin{figure*}
    \centering
    \includegraphics[width=\textwidth]{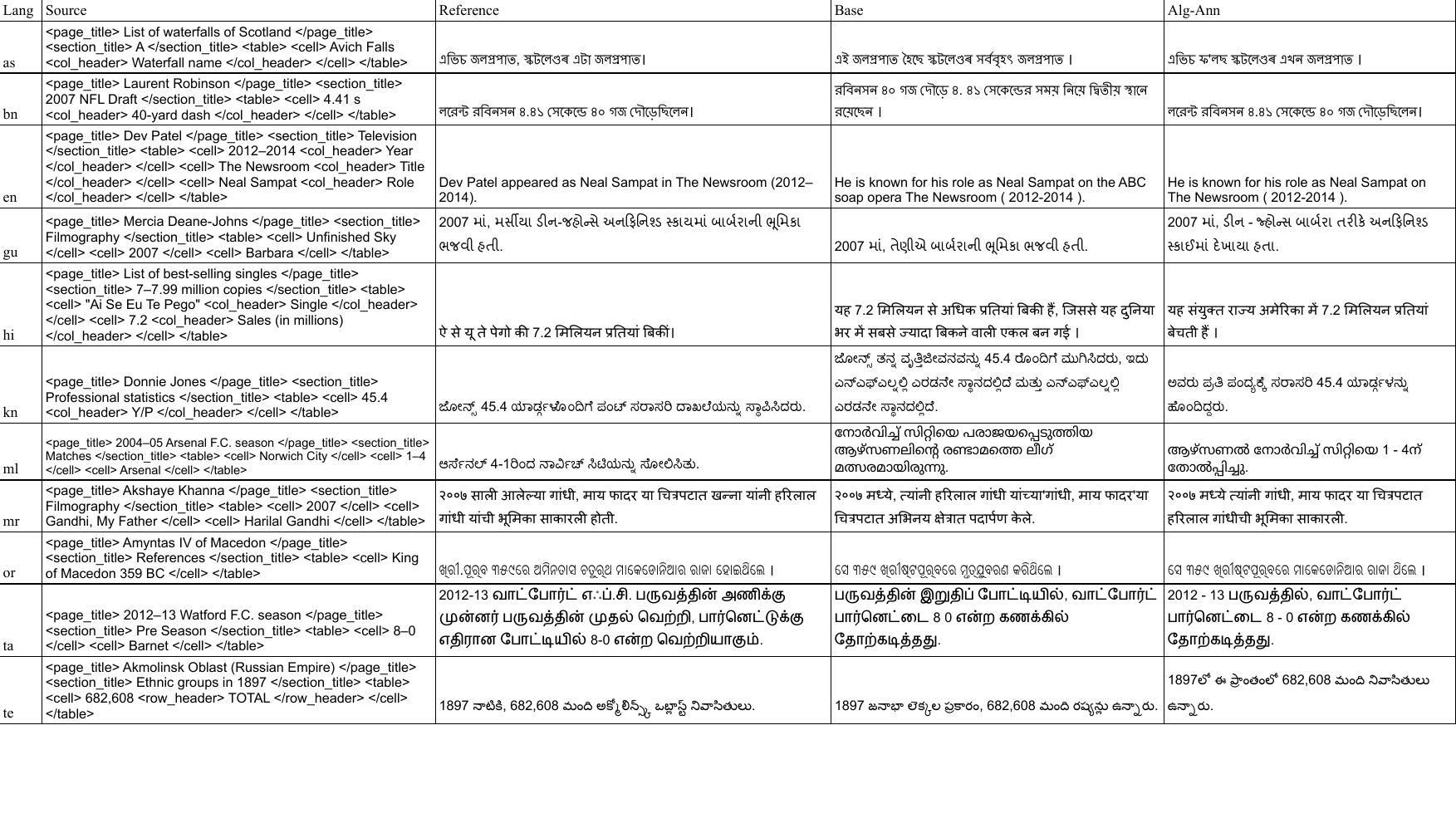}
    \captionof{table}{Some examples of generation from \textsc{xToTTo
    } using the best performing model compared to baseline model}
    \label{fig:totto_examples}
\end{figure*}
\end{document}